\documentclass[letterpaper, 10 pt, conference]{ieeeconf}  

\IEEEoverridecommandlockouts                              

\usepackage[T1]{fontenc}
\usepackage{url}
\usepackage{graphicx}
\usepackage{caption}
\newcommand{\eg}{\textit{e}.\textit{g}.}
\usepackage{amsmath}

\title{\LARGE \bf
 SiamEvent: Event-based Object Tracking via Edge-aware Similarity Learning with Siamese Networks
}

\author{Yujeong Chae, Lin Wang, and Kuk-Jin Yoon
\thanks{The authors are with the Visual Intelligence Laboratory, Department of Mechanical Engineering, KAIST, 34141 Daejeon, Republic of Korea (e-mail:
        {\tt\small yujeong@kaist.ac.kr, wanglin@kaist.ac.kr, kjyoon.kaist.ac.kr})}%
}

\begin{document}

\maketitle
\thispagestyle{empty}
\pagestyle{empty}

\begin{abstract}
Event cameras are novel sensors that perceive the per-pixel intensity changes and output asynchronous event streams, showing lots of advantages over traditional cameras, such as high dynamic range (HDR) and no motion blur. 
It has been shown that events alone can be used for object tracking by motion compensation or prediction. However, existing methods assume that the target always moves and is the stand-alone object. Moreover, they fail to 
track the stopped non-independent moving objects on fixed scenes.
In this paper, we propose a novel event-based object tracking framework, called SiamEvent, using Siamese networks via edge-aware similarity learning. Importantly, to find the part having the most similar edge structure of target, we propose to correlate the embedded events at two timestamps to compute the target edge similarity.
The Siamese network enables tracking arbitrary target edge by finding the part with the highest similarity score.
This extends the possibility of event-based object tracking applied not only for the independent stand-alone moving objects, but also for various settings of the camera and scenes.
In addition, target edge initialization and edge detector are also proposed to prevent SiamEvent from the drifting problem. Lastly, we built an open dataset including various synthetic and real scenes to train and evaluate SiamEvent.
The dataset is available at \url{https://github.com/yujeong-star/SiamEvent}.
Extensive experiments demonstrate that SiamEvent 
achieves up to 15\% tracking performance enhancement than the baselines on the real-world scenes and more robust tracking performance in the challenging HDR and motion blur conditions. 


\end{abstract}

\section{INTRODUCTION}

Event cameras are bio-inspired sensors that sense the per-pixel intensity changes asynchronously within a microsecond level and produce event streams encoding time, pixel location, and polarity (sign) of intensity changes. Compared with the traditional cameras, they show lots of advantages, such as high dynamic range (HDR), low latency and no motion blur.
Although events are sparse and mostly reflect the edges of scenes, it has been shown that an event camera alone can be applied to various vision tasks, such as depth and flow prediction~\cite{gehrig2019end,hidalgo2020learning,wang2021dual},  semantic segmentation~\cite{wang2021evdistill,alonso2019ev,wang2021dual} and object detection~\cite{cannici2019asynchronous,messikommer2020event}.

\begin{figure}[t]
\begin{center}
\includegraphics[width=\linewidth]{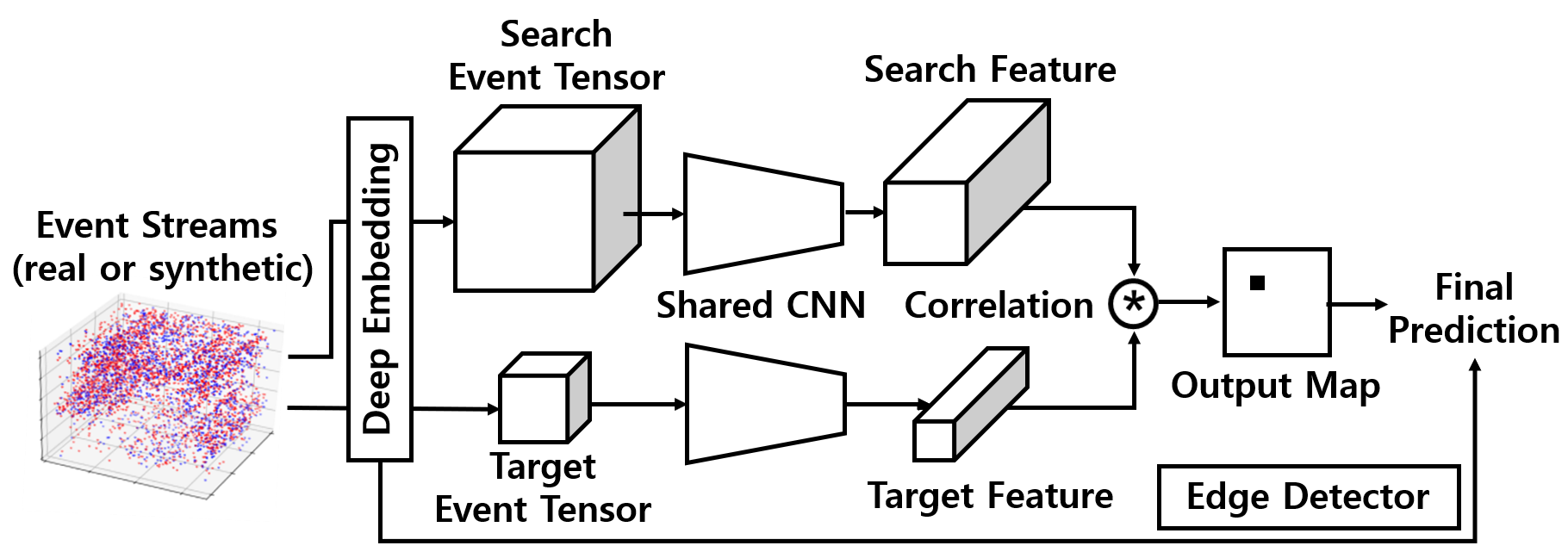}
\end{center}
    \vspace{-5pt}
    \captionsetup{font=small}
    \caption{SiamEvent correlates two features from the embedded search event tensor and target event tensor at two timestamps via a shared CNN to compute the target edge similarity. It then tracks the target edge by finding the part with the highest similarity score.}
    \vspace{-10pt}
\label{cover} 
\end{figure}

Object tracking is a task to estimate the state of the target object by utilizing the appearance and motion information of scenes. It has been shown that object tracking can benefit from the advantages of events' high temporal resolution. With this intuition, some event-based object tracking methods have been proposed in the literature \cite{mitrokhin, eventrnn, longterm, combine_fe, atsltd, robust}, among which the most representative idea is the motion compensation or prediction.
They could track the moving objects by estimating the motion parameter of a scene via minimizing the compensation error or maximizing the contrast of warped events and explicitly predicting the target state changes.
However, they assume that the target always moves and is the stand-alone object.
Therefore, they fail to track if the scene is fixed, the target is stopped, and there is the non-independent motion with background or non-target objects.
Moreover, they did not consider event distribution near the target. 
Although events triggered in the entire scene are chosen to compensate motion, it does not imply a meaningful movement of the target. The insufficient number of events that occurred on the target turns to be ignored as noise. 
Moreover, the target can not be re-tracked once it is hardly occluded. 

Unlike previous methods, we find that \textit{tracking with edges is not influenced by the type of object movements and enables following the non-independent moving target in static scenes. Moreover, the edges with sufficient number of events allow re-tracking the hardly occluded target and application to real world scenes}.
To this end, we propose a novel event-based object tracking framework, called \textbf{SiamEvent}, to efficiently track the target using the edge information provided by events in various settings of the camera and scenes. The core idea is to \textit{learn the edge-aware similarity using Siamese networks}. On one hand, to find the part having the most similar edge structure of the target, event streams at two timestamps are embedded while keeping the edge structure of the scene. Then, the two embedded events are fed to a shared feature extractor. The extracted features are finally correlated to compute the target edge similarity. 

On the other hand, the Siamese networks aim to learn the most similar part between two inputs and enables tracking arbitrary target edge by finding the part with highest similarity score. This extends the possibility of tracking applied \textit{not only for the independent, stand-alone moving objects but also for various settings of the camera and scenes}. In addition, target edge initialization and edge detector are also proposed to prevent SiamEvent from the drifting problem, occurring when objects similar to the target exist. Specifically, the target edge initialization generates a clear edge structure of the target and facilities training. The edge detector classifies whether the edge is distinctive and sufficient.

Previously, some datasets were made by filming the existing RGB image datasets in the monitor screen with an event camera \cite{monitor}. However, capturing data in this way failed to utilize high temporal resolution of events.
Other datasets \cite{mitrokhin, eventrnn} were captured only in the indoor scenes, thus limiting their applications.
Moreover, the sequences are small and simple, and some are not released to the public. It is possible to make synthetic datasets using the event simulator \cite{esim} based on the high frame rate image-based tracking dataset~\cite{nfs}. However, these datasets are applicable to solve real-world tracking problems. For this reason, we built an open event tracking dataset, including synthetic and real scenes. In particular, the real data were made by capturing the outdoor scenes with DAVIS346 \cite{davis346} event camera 
in various conditions, such as daytime, low light and fast motion scenes. Each sequence contains generic objects with various size. 

We evaluate our SiamEvent framework on the simulated and real datasets. Extensive experiments demonstrate that SiamEvent achieves 15\% tracking performance enhancement than the state-of-the-art (SoTA) image-based tracker \cite{siamfc} and event-based tracker \cite{mitrokhin} on the real-world scenes. 
Moreover, our method shows more robust tracking performance than the SoTA methods in the HDR and motion blur conditions. 

In summary, our main contributions are three folds. (I) We propose a novel event-based object tracking framework, called SiamEvent, using Siamese networks via edge-aware similarity learning, which allows tracking on various settings of the camera and scenes. Moreover, we propose target edge initialization and edge detector to prevent SiamEvent from the drifting problem.
(II) We built an event tracking dataset including various synthetic and real scenes, which will be open for public usage. (III) We conduct extensive experiments and demonstrate that SiamEvent achieves the state-of-the-art (SoTA) tracking performance on the real-world scenes and more robust tracking performance in the challenging HDR and motion blur conditions.

\section{Related Work}

\noindent \textbf{Event-based Object Tracking.}
Most conventional event-based object tracking frameworks were focused on moving object detection. The first attempt of tracking target using events was done by \cite{longterm} using tracking-learning-detection framework. Local search tracking and global search detection are jointly updated online. Later on, \cite{mitrokhin} proposed moving object detection framework with motion compensation by reducing the number of pixels of warped events and local spatial gradients of averaged timestamps of warped events. Then the moving objects are tracked with Kalman filter. \cite{atsltd} extracted objects proposals with contour-based detectors and set the proposal with largest IoU as the next target state.
Differently, our method tracks the object via edge-aware similarity learning. It considers the structure of target and enables tracking the non-independent moving object without additional detection phase.

To date, only a few methods have successfully applied deep learning to event-based object tracking.  
\cite{robust} used the pretrained classification network on the image data, \eg, VGG \cite{vgg}, to extract features from events 
and estimated target location by the correlation filter. However, as event data are in a different modality, such a method turns to extract less correct features, thus leading to less optimal tracking results. \cite{combine_fe} combined both frame and events and designed a two-layer convolutional neural network (CNN) to classify robot among four regions of interest based on the event data. 
Although \cite{eventrnn} 
employed the recurrent neural network (RNN) to estimate five object-level motion parameters from events, it can not handle the occluded scenes and becomes erroneous as it processes more events. Differently, our method can track the generic object even in challenging situations, \eg, occluded scenes, and prevent the drifting problem.

\noindent\textbf{Deep Learning on Event-based Vision.}
Deep learning was first used to solve basic classification and regression problems \cite{neil2016phased, maqueda2018event}. 
It was then applied to various visual perception tasks, \eg, object recognition \cite{rebecq2019high, bi2019graph}, detection \cite{hu2020learning,messikommer2020event},
semantic segmentation \cite{alonso2019ev,gehrig2020video,wang2021evdistill,wang2021dual}, flow and depth estimation \cite{tulyakov2019learning,gehrig2019end, stoffregen2020reducing,gallego2019focus,zhu2019unsupervised}. 

To bridge image and events, various works focused on low-level vision tasks. \cite{rebecq2019high, stoffregen2020reducing, mostafavilearning,wang2019event,wang2020eventsr, zhang2020learning,ParedesValls2020BackTE} focused on HDR and sharp image/video reconstruction from events. Moreover, some works focus on using events as guidance to  enhance image/video quality, such as image/video deblurring \cite{haoyu2020learning}, video frame interpolation and super-resolution \cite{ Tulyakov2021TimeLensEV,wang2020event}.
Differently, we propose a novel framework for object tracking, called SiamEvent, using Siamese networks via edge-aware similarity learning.

\noindent\textbf{Deep Learning on Frame-based Object Tracking.}
Some approaches tried to train a network to extract features offline and learn the target appearance online \cite{mdnet, goturn, ccot, eco}. Another line of research focused on similarity learning, which tracks the target by finding the most similar part between the target in the previous frame and current frame. \cite{siamfc} first applied Siamese networks in object tracking with fully-convolutional shared feature extractor. This allowed end-to-end training from feature extraction to correlation calculation, efficient generic object feature extraction and balanced tracking accuracy and speed. Following works explored additional information of target \cite{sa-siam}, region proposal network \cite{siamrpn},
deeper networks \cite{siamrpn++} and jointly learning object tracking and video object segmentation \cite{siammask}. In contrast, we propose an event-based object tracking framework, called SiamEvent, based on edge-aware similarity learning via Siamese networks. To find the part having the most similar edge structure of target, we propose to correlate the embedded events at two timestamps to compute the target edge similarity.

\begin{figure*}[t]
\begin{center}
\includegraphics[width=\linewidth]{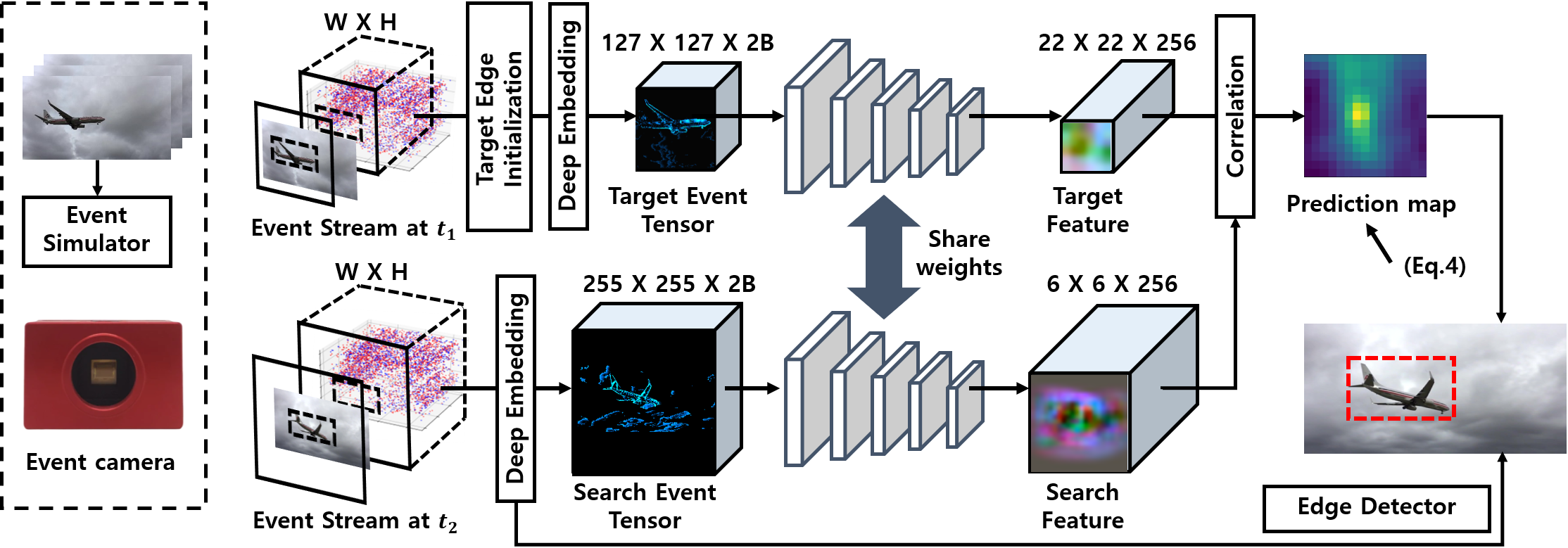}
\end{center}
    \vspace{-8pt}
    \captionsetup{font=small}
    \caption{Overview of the proposed SiamEvent framework, which comprises event embedding and Siamese networks via edge-similarity learning. Event streams at two timestamps are embedded and extracted features are correlated to compute target edge-similarity score.}
    \vspace{-10pt}
\label{main} 
\end{figure*}

\section{Method}

\subsection{Event representation}
We first describe the way to represent events, which can be fed to SiamEvent as inputs.  An event $e$ is interpreted as a tuple  ($\textbf{u}$, $t$, $p$), where $\textbf{u}$= ($x$, $y$) is the pixel coordinate, $t$ is the timestamp and $p$ is the polarity indicating the sign of intensity change. An event is triggered whenever a log-scale intensity change exceeds the threshold. 
Event spike tensor (EST) \cite{est} converts the asynchronous events to grid-like formats and allows to learn an event representation end-to-end for a given task. 
In an EST $S_{\pm}$,
a learnable kernel $k$ is applied to the event measurement field $f$. They are then put in a 4D voxel grid with the size of ($B$, 2, $H$, $W$), where $B$ is the number of bins, $H$ and $W$ are the height and width. 
$$
S_{\pm}[x_l, y_m, t_n] = (k \ * \  S_{\pm})(x_l, y_m, t_n) = 
$$
\begin{equation}
\label{eq:est}
\displaystyle\sum_{e_k \in \mathcal{E}_{\pm}} f_{\pm}(x_k, y_k, t_k)k(x_l-x_k, y_m-y_k, t_n-t_k).
\end{equation}
For $k$, we choose the trilinar kernel \cite{jaderberg2015spatial} for SiamEvent as it shows the best performance based on the experiments. 
As processing 4D tensor, \eg, 3D convolution, needs high computational cost and may degrade the test speed \cite{cui2019tensor}, 
we then reshape $S_{\pm}$ to a 3D tensor $S$ by stacking the voxel grids of two polarities.
In such a way, the tensor can be fed to our proposed framework to track an arbitrary target.


\subsection{Overview}
The overall framework of SiamEvent is depicted in Fig.~\ref{main}.  Events in the target-centered region $e_{TR}$ are selected and converted to EST while keeping the edge structure of the scene. The embedded events are fed to the shared feature extractor $F$ in the Siamese networks to extract features. We then find the correlations based on the extracted two features and finally compute the similarity score. 
The training is done by taking the event streams at two timestamps, $\{e\}_{t_1}^{t_2}$ and $\{e\}_{t_3}^{t_4}$, as inputs to our SiamEvent framework via the with edge-aware similarity learning (Sec.~\ref{edge_aware_learning}). Online tracking process takes two continuous event streams, $\{e\}_{t_1}^{t_2}$ and $\{e\}_{t_2}^{t_3}$, as the inputs and sets the target state at the part with highest edge similarity score. 
Moreover, to prevent tracker drifting problem, we propose the target edge initialization (Sec.~\ref{target_edge}), which allows generating a clear edge structure of the target. The edge detector are then proposed to classify whether the edge is sufficient.

\subsection{Edge-aware Similarity Learning via Siamese Networks}
\label{edge_aware_learning}
We first describe the structure of the Siamese networks, which are two-stream networks sharing weights but taking different inputs to measure the edge similarity between them.
Various architectures of networks as chosen as the shared feature extractor in Siamese networks \cite{siamfc, sa-siam, siamrpn, siamrpn++, siammask}. 
As events are sparse and contain lots of empty grids, very deep networks are hard to train due to the vanishing gradient problem. 
Therefore, we adopt AlexNet \cite{alexnet} as the shared feature extractor. Differently, we modify the structure by adding the batch normalization layers after every convolutional layer and removing the fully connected layers. Moreover, the number of channels in first convolutional layer is set to be twice of $B$. 

\begin{figure}[t]
\begin{center}
\includegraphics[width=.95\linewidth]{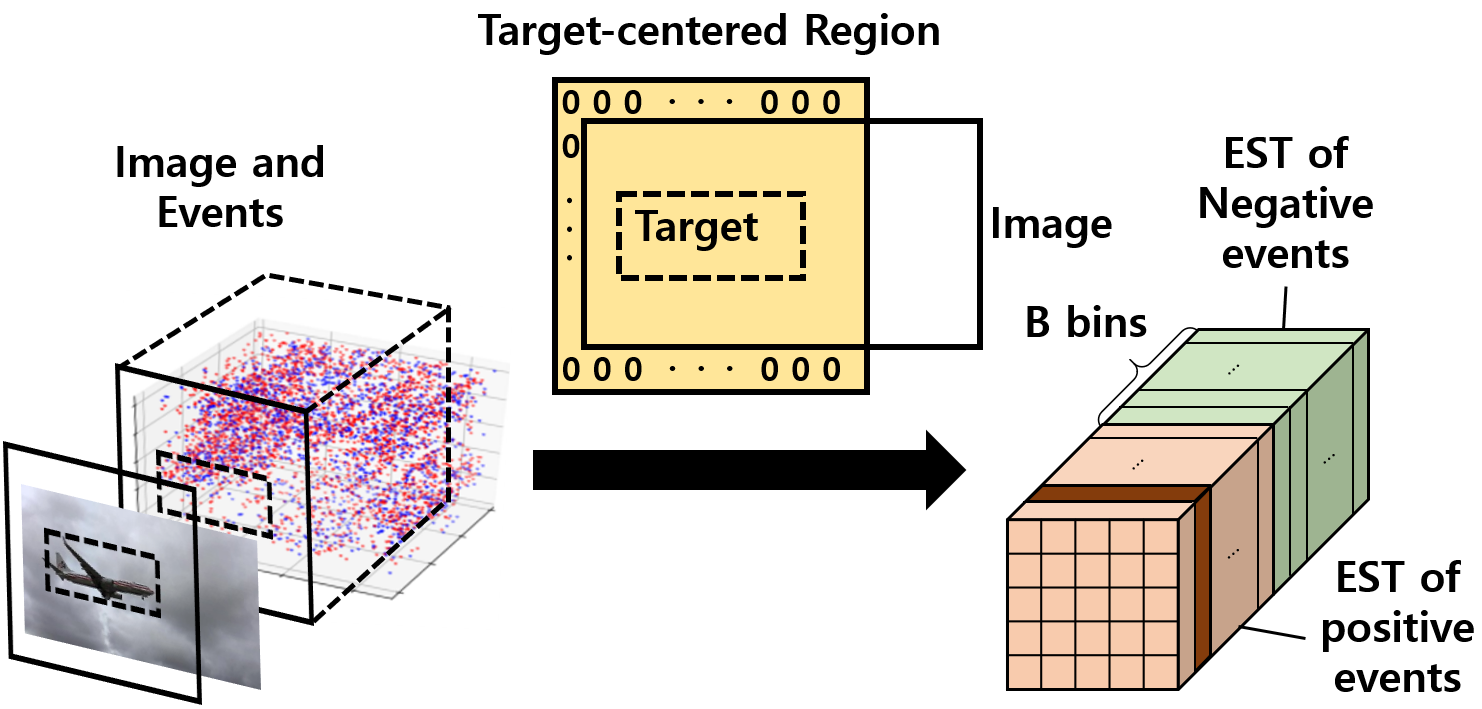}
\end{center}
    \vspace{-8pt}
    \captionsetup{font=small}
    \caption{Event selection for similarity learning. The events whose coordinates are within the target-centered region are selected for learning a representation for SiamEevnt.}
    \vspace{-10pt}
\label{target_centered_event} 
\end{figure}

The second crucial step is how to select events to be used for
similarity learning. 
Inspired by \cite{siamfc}, the events whose spatial coordinates are within the target-centered region $e_{TR}$, are selected for learning an event representation, as shown in Fig.~\ref{target_centered_event}.
If the target-centered region exceeds the original spatial resolution, the corresponding parts are filled with zeros. 
The size of target-centered region is set as $(\frac{1}{2} h + \frac{3}{2}w) \times (\frac{1}{2}w + \frac{3}{2}h)$, where $w$ and $h$ are the width and height of the target. The selected events are then converted to an EST, $S{e_{TR}}$, which enables jointly learning the event representation and tracking network. 

More specifically, the event conversion is done at two different timestamps for similarity learning. The events with the earlier timestamp $\{e_{TR}\}_{t_1}^{t_2}$ are resized to 127 $\times$ 127 with bilinear interpolation and set as the target event tensor. The events  with the later timestamp $\{e_{TR}\}_{t_3}^{t_4}$ are resized to 255 $\times$ 255 with the bilinear interpolation and set as the search event tensor. The event streams at two timestamps are embedded to $S(\{e_{TR}\}_{t_1}^{t_2})$ and $S(\{e_{TR}\}_{t_3}^{t_4})$ while keeping the edge structure. This allows to find the part having the most similar edge structure of the target in the search space.

With the events being selected and converted, the last step is how to compute the edge similarity.
To compute the similarity between the target and search event tensors, we first feed them to the shared feature extractor $F$. As such, we can obtain a target feature $F(S(\{e_{TR}\}_{t_1}^{t_2}))$ and a search feature $F(S(\{e_{TR}\}_{t_3}^{t_4}))$, respectively, as shown in Fig.~\ref{main}. 
These two features are correlated to compute target edge similarity with correlation $\phi$, and prediction map $P$ is formulated as follows:
\begin{equation}
\label{eq:outputmap}
    P = \phi\{F(S(\{e_{TR}\}_{t_1}^{t_2})), F(S(\{e_{TR}\}_{t_3}^{t_4}))\}.
\end{equation}

As the overall structure is fully-convolutional networks without padding operation, the computed target edge similarity should have the highest value in the center of the correlation map. 
Therefore, the label of correlation map $L$ is set to have one near the center and zero otherwise as follows: 
\begin{equation}
\label{eq:labelmap}
    L = 
    \begin{cases}
        1,& \text{if } ||(x, y)-c|| \leq R,\\
        0,              & \text{otherwise,}
    \end{cases} \\
\end{equation}
where $c$ is the center of prediction map and $R$ is distance threshold.
The binary cross entropy loss \cite{siamfc} is used to compute the similarity between the prediction map and label map, which is formulated as:
\begin{equation}
\label{eq:bceloss}
    Loss(P, L) = {\frac{1}{|A|}\sum\limits_{(x, y)\in A} (log(1 + exp(-P(x, y)L(x, y)))}.
\end{equation}
With edge-aware similarity learning, SiamEvent can be applied 
to various settings of the camera and scenes.

\begin{figure}[t]
\begin{center}
\includegraphics[width=.98\linewidth]{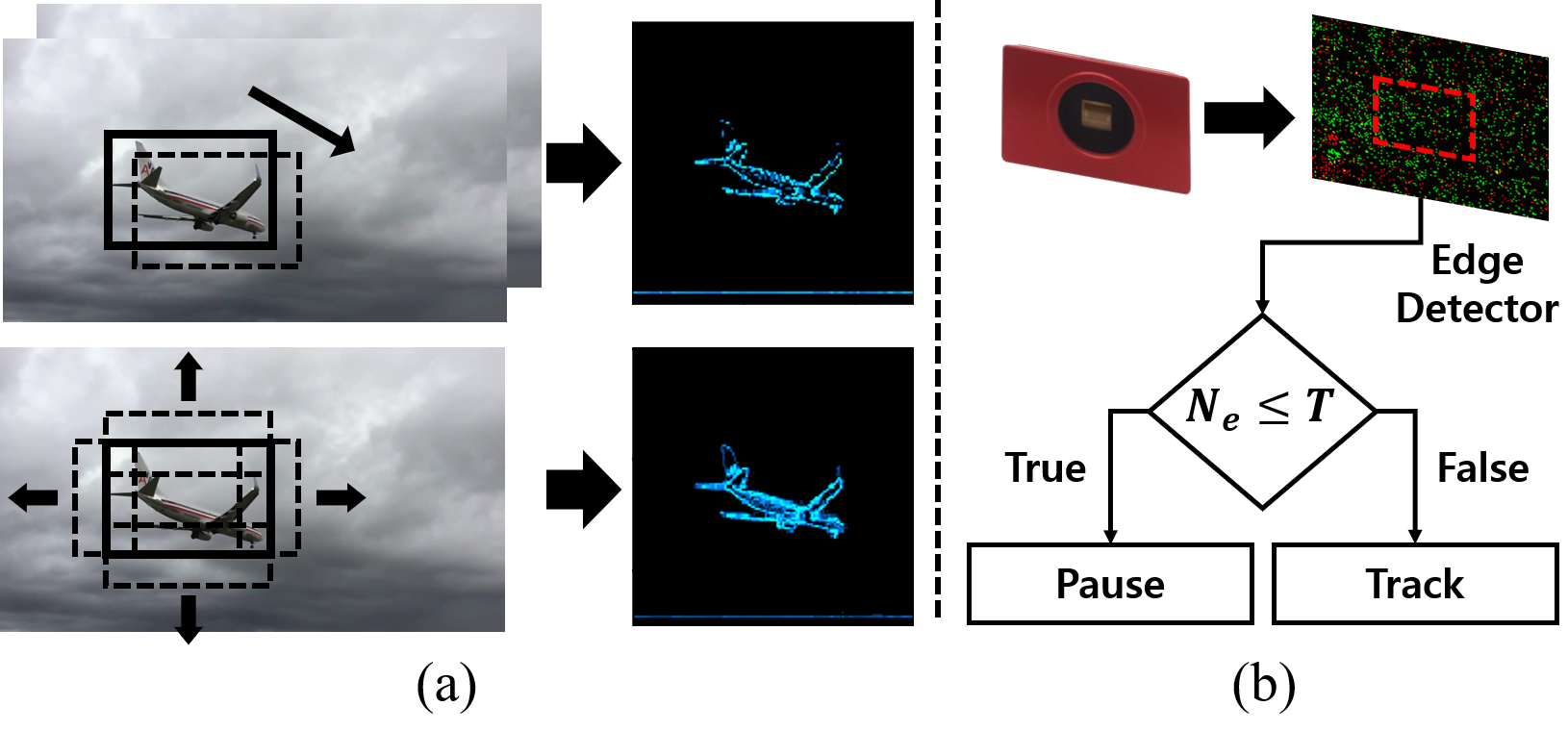}
\end{center}
    \vspace{-12pt}
    \captionsetup{font=small}
    \caption{(a) Target edge initialization allows complete and clear edges of the scene (b) Edge detector classifies whether the edges of embedded events are distinctive and sufficient for tracking.}
    \vspace{-8pt}
\label{init} 
\end{figure}

\subsection{Target Edge Initialization and Edge Detector}
\label{target_edge}
Additionally, we propose target edge initialization to prevent our method from the drifting problem. We also propose edge detector to enable our method work in the static scene. 

As events are triggered in the scene based on moving direction of the camera or objects, incomplete and partial edges of events can degrade the tracking performance. Therefore, we propose the target edge initialization. It generates complete target edges to prevent the drifting problem that occurs when the tracker confuses the target with similar objects.
For synthetic events, target edge initialization is done by moving the virtual camera of the simulator along the first image frame of the sequence. It moves horizontally and vertically with the length of 0.01 multiplies the sum of target width and height, allowing to generate complete and clear edges of a scene. 
Compared with the one by directly using event streams in earlier timestamps as the target, we find that the proposed target edge initialization generates clear and complete edges, as shown in Fig.~\ref{init}(a). 
For real events, as the sequences are taken in the low light and motion blur conditions. Therefore, it is impossible to simulate events using the active pixel sensor (APS) frames (provided by the DAVIS 346 camera). To tackle this problem, we give little linear motion horizontally and vertically to the event camera to fully utilize the edge information of a scene.



To enable tracking in the static scenes, we also propose a simple yet effective edge detector to prevent tracking failure. The reason is that, in static scene, frame-based cameras have outputs; however, event cameras do not have any outputs. The edge detector is a classifier that discerns whether the edges of embedded events are distinct and sufficient for similarity learning.
In particular, as shown in Fig.~\ref{init}(b), the edge detector pauses the tracker to skip tracking ($O(N_e)=1$) if the number of events $N_e$, occurred in the target-centered region, doesn't exceed the threshold $T$. The principle is formulated in Eq.~\ref{eq:edge_det}. Here, $O$ is the output of the edge detector based on the event number $N_e$. 
As the target-centered region depends on the target size, we set the threshold $T$ as 0.05 $\times$ $(\frac{1}{2} h + \frac{3}{2}w) \times (\frac{1}{2}w + \frac{3}{2}h)$ to make it proportional to the area of target-centered region.

\begin{equation}
\label{eq:edge_det}
O(N_e) =
    \begin{cases}
        Skip, & \text{if } N_e \leq T,\\
        Continue, & \text{otherwise.}
    \end{cases} \\
\end{equation}

When the tracker is paused, the operations, \eg, feature extraction and correlation computation, are terminated. As such, the target state is predicted same as the previous one. Therefore, the edge detector removes unnecessary computations and increases the tracking efficiency in the static scenes near the target.

\begin{figure}[t]
\begin{center}
\includegraphics[width=\linewidth]{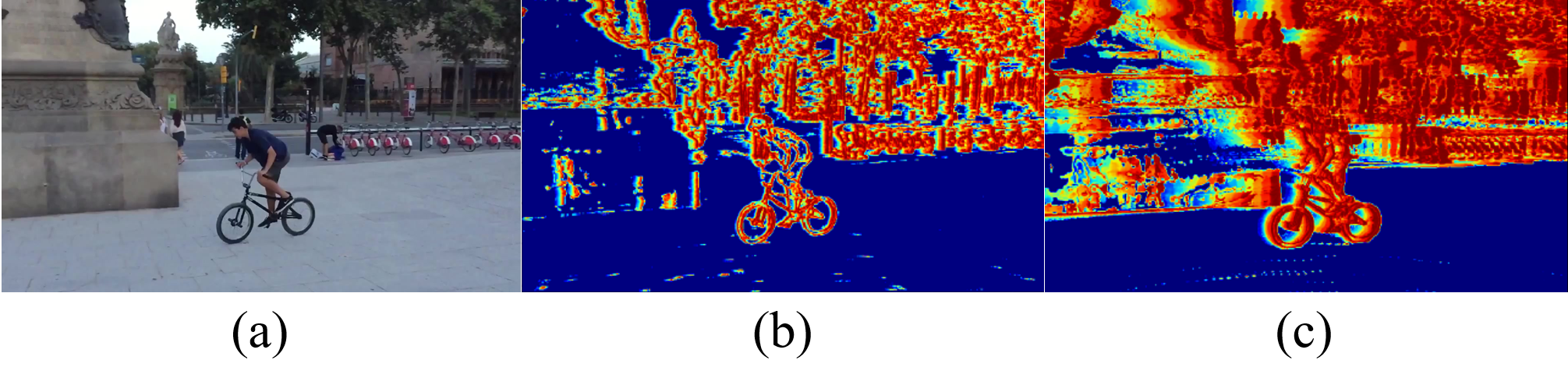}
\end{center}
    \vspace{-8pt}
    \captionsetup{font=small}
    \vspace{-8pt}
    \caption{(a) RGB image of NFS \cite{nfs}, simulated events between two images by putting recent timestamps when (b) the frame rate is 240fps (c) the frame rate is 30fps. ESIM can generate more realistic events when the time interval between two frames is closer.}
    \vspace{-5pt}
\label{esim240} 
\end{figure}

\begin{figure}[t]
\begin{center}
\includegraphics[width=\linewidth]{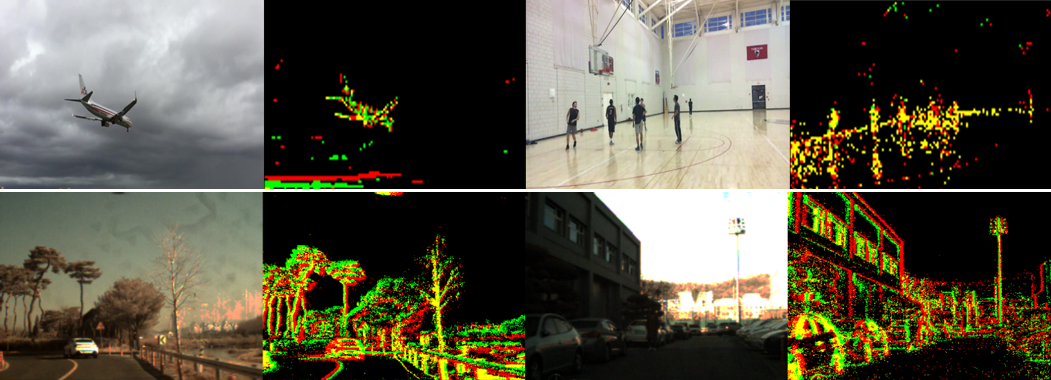}
\end{center}
    \vspace{-5pt}
    \captionsetup{font=small}
    \caption{Example images (1st and 3rd columns) and visualized events (2nd and 4th columns) with stacking along polarities of synthetic (1st row) and real-world (2nd row) event datasets.}
    \vspace{-10pt}
\label{data_event} 
\end{figure}

\begin{figure*}[t!]
\begin{center}
\includegraphics[width=\linewidth]{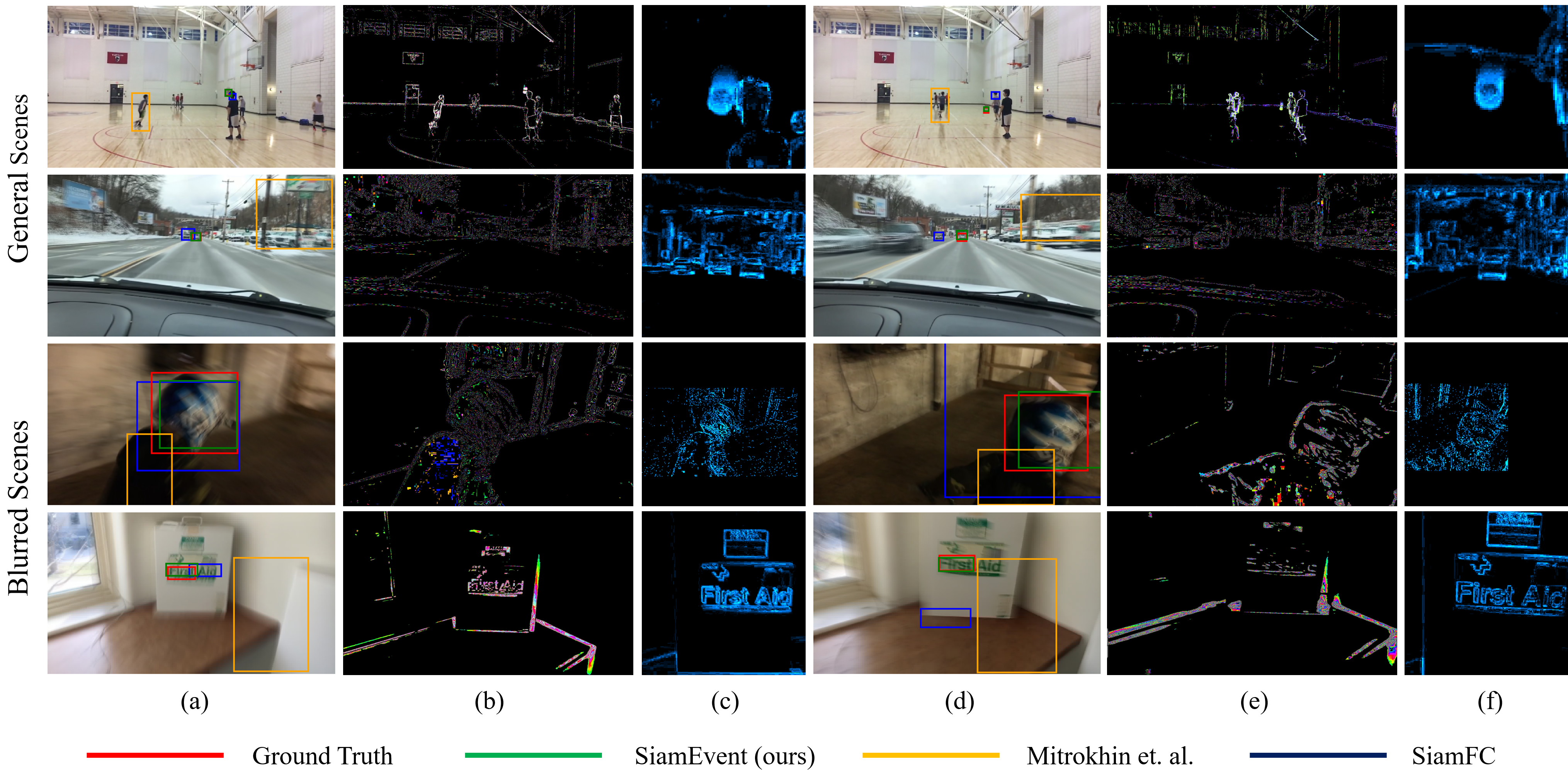}
\end{center}
    \vspace{-8pt}
    \captionsetup{font=small}
    \caption{Experimental results on the synthetic dataset at two timestamps $t_1$ and $t_2$. (a) tracking results of our SiamEvent visualized on color frames at time $t_1$, (b) events from \cite{mitrokhin} at $t_1$, (c) search event tensor at $t_1$, (d) tracking results visualized on color frames at time $t_2$, (e) events from \cite{mitrokhin} at $t_2$, (f) search event tensor at $t_2$.}
    \vspace{-6pt}
\label{synqualmore5_3} 
\end{figure*}

\subsection{Dataset Preparation}
\label{dataset_sec}
\subsubsection{Simulated Tracking Dataset}
\label{syndataset}
Previously, the simulated datasets were made by filming the existing image datasets in the monitor screen with an event camera \cite{monitor}. However, capturing data in this way fails to utilize high temporal resolution of events, and the data have lower resolution and smaller number of sequences than the image datasets. 
To this end, we have built a simulated large-scale event-based tracking dataset using the existing image-based tracking dataset NFS \cite{nfs} via the event simulator (ESIM) \cite{esim}. the positive and negative constant thresholds of ESIM are set as 0.15.  The bounding box labels of the NFS dataset are directly used as the ground truth (GT) labels.  ESIM takes image sequences as inputs and generates synthetic events between the images with the corresponding timestamps. 
NFS dataset is the only tracking dataset that provides high frame rate image sequences. Moreover, the images are in a high spatial resolution, \eg, 1280 $\times$ 720. It includes 100 video sequences and 38$K$ bounding boxes with 33 target classes. It also has two versions of frame rate: 30fps and 240fps.
The latter one is used for ESIM that to make higher quality dataset, as shown in Fig.~\ref{esim240}.  
Moreover, the 1st row of Fig.~\ref{data_event} shows some examples of synthetic event dataset.  


\subsubsection{Real-world Tracking Dataset}
\label{realdataset}
Several real-world tracking datasets \cite{mitrokhin, eventrnn} have been proposed in the literature. However, the sequences are captured only in the indoor scenes, thus limiting their applications. Moreover, the sequences are relatively less qualitative and quantitative. Importantly, most of them are not released to the public. 
Therefore, we have built a new real-world event dataset by capturing the outdoor scenes with a DAVIS346 \cite{davis346} event camera with a spatial resolution of 260 $\times$ 346.  Overall, eight sequences are captured in various conditions, \eg, daytime, low light and fast motion scenes. Moreover, the dataset was captured with various settings of the camera and scenes, \eg, fixed camera with the stand-alone moving object, the moving camera with fixed scenes. 
We considered four main conditions in the dataset: general, motion blur, over-exposed and under-exposed conditions. The target size and classes are also varied for same reason. 
Bounding boxes of the target are annotated manually. The 2nd row of Fig.~\ref{data_event} shows some examples of real events. Our simulated and real-world datasets will be released to incite more research by the community.

\begin{table}[t!]

\caption{Comparison with SiamFC \cite{siamfc} and \cite{mitrokhin} on the synthetic event dataset. \textbf{Bold} numbers indicate our results.}
\vspace{-10pt}
\label{table:syn}
\begin{center}
\resizebox{\linewidth}{!}{
\begin{tabular} {llccc}
\hline\hline
 & Dataset & \begin{tabular}{@{}c@{}}Success \\ score\end{tabular} & \begin{tabular}{@{}c@{}}Precision \\ score\end{tabular} & \begin{tabular}{@{}c@{}}Success \\ rate\end{tabular} \\
\hline
SiamFC \cite{siamfc}&NFS& 0.565 & 0.660 & 0.692\\
SiamFC \cite{siamfc}& 8-merged NFS& 0.475 & 0.528 & 0.557\\
SiamFC \cite{siamfc}& 16-merged NFS& 0.382 & 0.390 & 0.424\\
Mitrokhin et. al. \cite{mitrokhin} & NFS + ESIM & 0.137& 0.121 & 0.135\\
SiamEvent (ours) & NFS + ESIM & \textbf{0.404} & \textbf{0.454} & \textbf{0.478}\\
\hline\hline
\end{tabular}}
\end{center}
\vspace{-15pt}
\end{table}

\begin{table*}[t!]
\captionsetup{font=small}
\caption{Comparison of SiamEvent with the SoTA frame-based tracking method \cite{siamfc} and event-based tracking method \cite{mitrokhin} on our real dataset. The success rate, prediction rate and precision rate are measured for each sequence. \textbf{Bold} numbers indicate our results.}
\vspace{-10pt}
\label{table:real}
\begin{center}
\resizebox{\linewidth}{!}{
\begin{tabular} {llccccccccc}
\hline\hline
 & Sequence & "arrow" & "car1"& "car2"&"car3"&"car4"&"gs25"&"person1"&"person2"&Overall \\
 & Description & blurred&strong light&general&general&strong light&blurred&dark&general&- \\
\hline
& Success score& 0.769&0.357&0.742&0.589&0.316&0.476&0.190&0.756&0.524 \\
SiamFC \cite{siamfc} & Precision score&1.000&0.833&1.000&1.000&0.363&0.857&0.200&1.000&0.781 \\
& Success rate&1.000&0.166&1.000&0.769&0.272&0.571&0.200&1.000&0.622\\
\hline
& Success score&0.531&0.698&0.286&0.066&0.515&0.048&0.629&0.008&0.348\\
Mitrokhin et. al. \cite{mitrokhin}& Precision score& 1.000&1.000&0.550&0.154&0.636&0.000&0.800&0.000&0.518\\
&Sucess rate & 0.666&1.000&0.200&0.000&0.636&0.000&0.800&0.000&0.291\\
\hline
& Success score& 0.738&0.539&0.709&0.523&0.372&0.625&0.733&0.658&\textbf{0.612}\\
SiamEvent (ours) &Precision score&1.000&1.000&1.000&1.000&0.454&1.000&1.000&1.000&\textbf{0.931}\\
& Success rate&1.000&0.500&1.000&0.461&0.181&0.714&1.000&0.833&\textbf{0.711}\\
\hline\hline
\end{tabular}}
\end{center}
\vspace{-12pt}
\end{table*}

\section{Experiments}

\subsection{Implementation and Training Details}
In this section, we present the experimental results on the datasets described in Sec.\ref{dataset_sec}.
For the synthetic dataset, 80\% of the sequences were randomly chosen for training and the others for test.  The number of bins $B$ in reshaped EST and distance threshold $R$ in Eq.~\ref{eq:outputmap} are set as 9 and 3, respectively.
Our method is implemented with PyTorch using one Titan RTX 2080 Ti GPU. We initialized the network with dynamic learning rate and trained our method for 50 epochs. We set the batch size of 1 as the number of events in each batch differs. SGD optimizer with the weight decay of $5e^{-4}$ and momentum of 0.9 is used. Cosine window with 0.176 window influence is applied for distractor suppression.

We use the following metrics to evaluate the tracking performance. The precision is the ratio of frames that has smaller distance than threshold between the centers of prediction and ground truth (GT). As threshold changes from 0 to 1, the precision curve is made with the threshold as the x-axis and the precision as the y-axis. The area under precision curve is set as the precision score. The principle of success score is same as the precision score, but with the overlap area instead of center distance \cite{siamfc, siamrpn}. Success rate is the ratio of frames that has overlap ratio between the prediction and GT (larger than 0.5) \cite{mitrokhin, eventrnn, atsltd}. The running speed is around 25fps (event aggregation and target tracking).

\subsection{Evaluation on Synthetic Dataset}
The quantitative and qualitative results are shown in Table.~\ref{table:syn} and Fig.~\ref{synqualmore5_3}. We compare our method with the image-based tracker SiamFC \cite{siamfc} and the event-based tracker \cite{mitrokhin}. \textit{Please be noted that we could not compare with \cite{eventrnn,robust} as the datasets and codes were not publicly available}.
Moreover, we compare with SiamFC on the motion blur condition. SiamFC is tested on the blur NFS dataset, which is made by averaging consecutive $N$ 240-fps frames, inspired by~\cite{deblurgan}. We name it as $N$-merged NFS dataset.


\begin{figure*}[t]
\begin{center}
\includegraphics[width=\linewidth,height=9.5cm]{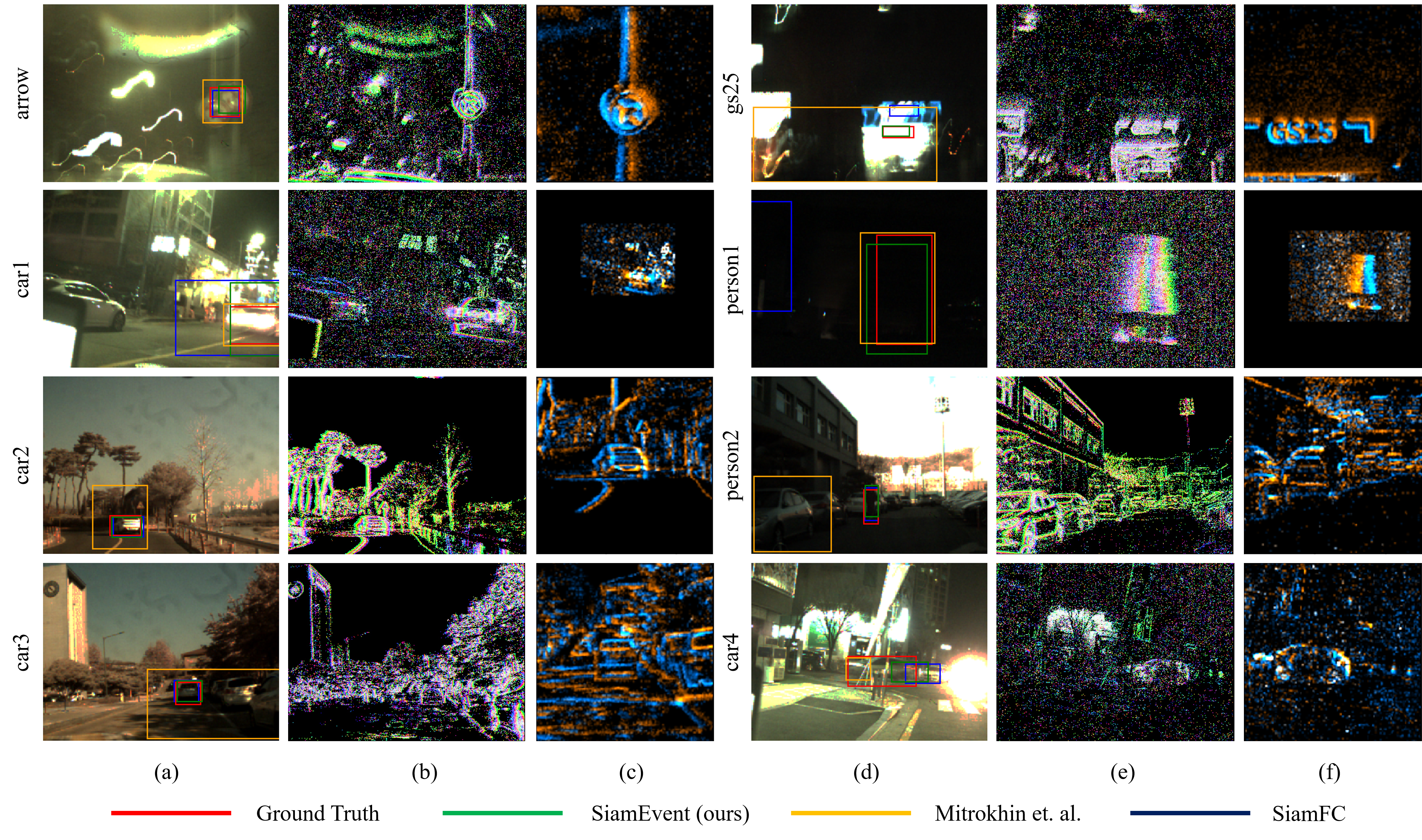}
\end{center}
    \vspace{-10pt}
    \captionsetup{font=small}
    \caption{Experimental results on the real event dataset. (a) and (d) tracking results of our SiamEvent visualized on the APS frames; (b) and (e) events from \cite{mitrokhin}; (c) and (f) search event tensors.}
    \vspace{-8pt}
\label{realqualres} 
\end{figure*}

Compared with the event-based tacker \cite{mitrokhin}, our SiamEvent significantly surpasses it by around 27\% and 33\% increase of success and precision score. As \cite{mitrokhin} can only track the stand-alone and isolated object, it could not properly divide the target from the background under various scenes. Moreover, the target can not be re-tracked when occlusion occures. 

Our SiamEvent also shows better performance than the image-based tracker \cite{siamfc} on the challenging blurred scenes. We get 2\%, 6\%, 5\% increase of the success score, precision score, and success rate on the 16-merged NFS dataset, respectively. Qualitatively, as shown in Fig.~\ref{synqualmore5_3}, SiamFC confuses the target with other objects and can not estimate the accurate bounding box of the target. However, our SiamEvent tracks the target without drifting and estimates tight bounding box with the target edges. 

\subsection{Evaluation on Real Event Dataset}

We now present the experimental results on the real-world dataset, especially under the \textit{low dynamic range and motion blur} conditions. The quantitative and qualitative results are shown in Table.~\ref{table:real} and Fig.~\ref{realqualres}. 

Our SiamEvent surpasses SiamFC and \cite{mitrokhin} by a noticable margin with around 10\% to 15\% increase regrading the three metrics. In particular, SiamEvent outperforms SiamFC in the blurred, over-exposed and under-exposed conditions. For instance, SiamEvent could track the target that is occluded before, in contrast to SiamFC, in the "car4" sequence. Compared with the event-based tracker \cite{mitrokhin}, SiamEvent also surpasses it by a large margin. The results indicate that our method can better utilize edge information of events to track the target. Moreover, our SiamEvent tracker can better track the target in the fixed scene with a moving camera, \eg, in the "arrow" and "gs25" sequences. In the general condition, SiamEvent shows much higher performance than \cite{mitrokhin}. 

In summary, our SiamEvent framework using similarity learning via Siamese Networks achieves the state-of-the-art performance on the synthetic and real-world data. The tracking results demonstrated that our method fully exploits the edges information from events and better track the non-independent moving and stand-alone objects in general scenes, as well as the HDR and motion blur scenes.

\subsection{Ablation Studies and Analyses}

\noindent \textbf{Event Embedding vs. Tracking Performance.}
 Event representation is an important factor for the tracking performance.
Although we mainly used EST \cite{est} for event representation for SiamEvent, we compared with other three methods. As shown in Table~\ref{table:embedmethod}, the one-channel representation is the worst as it uses the least amount of information of the scene. The two-channel representation shows better performance than one-channel representation. The reason is that it leverages the polarity information and suppresses the noise in the events. EST achieves the best tracking accuracy as it can focus on the timestamps of recently occurred events without directly counting the entire event streams. It allows to learn an end-to-end event representation for target tracking.

\begin{table}[t!]
\caption{Ablation study of embedding method on synthetic event dataset.}
\vspace{-8pt}
\label{table:embedmethod}
\begin{center}
\resizebox{\linewidth}{!}{
\begin{tabular} {lccc}
\hline\hline
 Embedding method& Success score & Precision score & Success rate \\
\cline{1-4}
One-channel image & 0.345 & 0.362 & 0.371\\
Two-channel image & 0.355 & 0.370 & 0.398\\
Two-channel voxel & 0.377 & 0.411 & 0.442\\
EST \cite{est} & \textbf{0.404} & \textbf{0.454} & \textbf{0.478} \\
\hline\hline
\end{tabular}}
\end{center}
\vspace{-14pt}
\end{table}

\vspace{2pt}
\noindent \textbf{Target Edge Initialization.} 
We look into the effect of target edge initialization. As shown in Table.~\ref{table:static}, without target edge initialization, the performance drops from 0.404 to 0.389 in success score, from 0.454 to 0.434 in the precision score and from 0.478 to 0.441 in success rate. The results indicate that target edge initialization is important for the success of robust tracking. 

\begin{table}[t!]
\caption{Ablation study of target edge initialization and edge detector on synthetic event dataset.}
\vspace{-8pt}
\label{table:static}
\begin{center}
\resizebox{\linewidth}{!}{
\begin{tabular} {lccc}
\hline\hline
& Success score & Precision score & Success rate \\
\cline{1-4}
w/o initialization & 0.389 & 0.434 & 0.441\\
w/o edge detector & 0.285 & 0.342 & 0.303 \\
SiamEvent & \textbf{0.404} & \textbf{0.454} & \textbf{0.478}\\
\hline\hline
\end{tabular}}
\end{center}
\vspace{-15pt}
\end{table}

\vspace{2pt}
\noindent \textbf{Edge Detector.} Edge detector helps tracker to track the target in static scenes. Without edge detector, there exists performance drop, as shown in Table.~\ref{table:static}. When other settings are fixed, the tracking accuracy drops from 0.404 to 0.285 in precision score, from 0.454 to 0.342 in success score and from 0.478 to 0.303 in success rate. The results show that edge detector handles the static scenes and prevents the tracker from the drifting problem. 

\noindent \textbf{Speed Analysis.} Tracking on the real events operates at 25fps (including event embedding). When the events are already embedded, the tracking speed increases to 120fps. Tracking on the synthetic events has additional step for event simulation with ESIM \cite{esim}. If event simulation and tracking run together, tracker operates at 10fps. Moreover, using the edge detector changes the speed from 37fps to 25fps.

\begin{figure}[t!]
\begin{center}
\includegraphics[width=\linewidth]{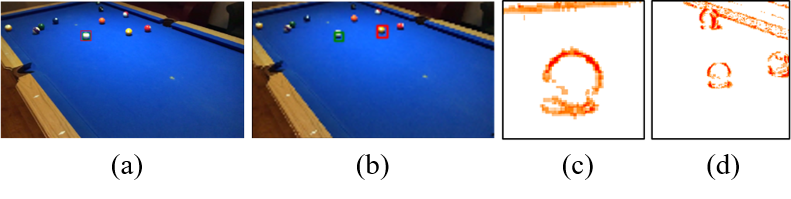}
\end{center}
    \vspace{-10pt}
    \captionsetup{font=small}
    \caption{Failure case on the synthetic event dataset. Green box indicates the GT, (a) tracking result of SiamFC in red (b) tracking result of SiamEvent in red, (c) target feature (d) search feature.}
    \vspace{-10pt}
\label{synqualresult1} 
\end{figure}

\noindent \textbf{Failure cases.} 
SiamEvent failed to track the target when there are the same objects with different colors and the target is unclear and featureless. As shown in Fig.~\ref{synqualresult1}, SiamEvent is confused by the white ball (target) with the yellow ball while SiamFC successfully tracks the white ball. If the target has its own appearance, as shown in second row of Fig.~\ref{synqualresult1}, SiamEvent can track the target in such a situation.

\section{CONCLUSIONS}
In this paper, we have proposed a novel yet efficient framework, called SiamEvent, empowered by the Siamese networks via edge-aware similarity learning, for event-based object tracking. To find the part having the most similar edge structure of target, we proposed to probe the correlations between the extracted features of the embedded events at two timestamps to compute the edge similarity score. We have also built an open data including simulated and real-world events. Extensive experiments have demonstrated the effectiveness of EventSiam on the challenging HDR and motion blur conditions. 










\bibliographystyle{ieeetran}
\bibliography{IEEEabrv, egbib}

\end{document}